\documentclass[letterpaper, 10 pt, conference]{ieeeconf_2}
\IEEEoverridecommandlockouts

\usepackage{cite}
\usepackage{amsmath,amssymb,amsfonts}
\usepackage{algorithmic}
\usepackage{graphicx}
\usepackage{textcomp}
\usepackage{booktabs}
\usepackage{xcolor}
\usepackage{multirow}
\usepackage{url}
\usepackage{hyperref}
\usepackage[font=normalfont,labelfont=bf]{caption}
\def\BibTeX{{\rm B\kern-.05em{\sc i\kern-.025em b}\kern-.08em
    T\kern-.1667em\lower.7ex\hbox{E}\kern-.125emX}}
\begin{document}

\title{FLaRA: Predicting Future Latent Representations for Accident Anticipation}

\author{
 	\parbox{\textwidth}{%
 		\centering
 		Lorenzo Caselli$^{1}$, Tomaso Trinci$^{2}$, Tommaso Bianconcini$^{2}$, Simone Magistri$^{1}$, \\ Leonardo Taccari$^{2}$, Francesco Sambo$^{2}$, Andrew D. Bagdanov$^{1}$%
 	}%
 	\thanks{$^{1}$Department of Information Engineering, University of Florence, Italy.}
 		  \thanks{Email: {\tt name.surname@unifi.it}}
        \thanks{$^{2}$Verizon Connect, Florence, Italy.}
 		  \thanks{Email: {\tt name.surname@verizonconnect.com}}
        \thanks{\copyright 2026 IEEE. Personal use of this material is permitted. Permission from IEEE must be obtained for all other uses, in any current or future media.}
    }

\maketitle

\begin{abstract}
Anticipating traffic accidents from dashcam videos is a critical challenge in intelligent transportation systems. Existing methods typically map visual context directly to a collision probability without explicitly modeling the future evolution of the driving scene. In this paper we propose FLaRA (Predicting Future Latent Representations for Accident Anticipation), a novel predictive architecture that shifts this paradigm by forecasting future latent representations for accident anticipation. Building upon the Video Joint-Embedding Predictive Architecture (V-JEPA2), our model conditions a predictor network on observed context frames to predict the forthcoming latent features of the scene. A classifier then operates on these predicted future representations rather than only on past observations. To ensure these forecasts remain grounded in realistic future dynamics, we introduce a joint training objective that simultaneously optimizes an auxiliary feature-level reconstruction loss and a cross-entropy classification loss. Extensive evaluations on the Nexar dataset, alongside cross-domain validations on the DAD, DADA-2000, and DoTA benchmarks, demonstrate that our approach achieves state-of-the-art performance while maintaining realistic early warning capabilities. The code is publicly available at: \small{\href{https://github.com/LoreCase073/FLaRA}{\url{https://github.com/LoreCase073/FLaRA}}}.
\end{abstract}

\section{Introduction}
\label{sec:introduction}

\begin{figure*}[t]
\centering
\centerline{\includegraphics[width=0.75\textwidth]{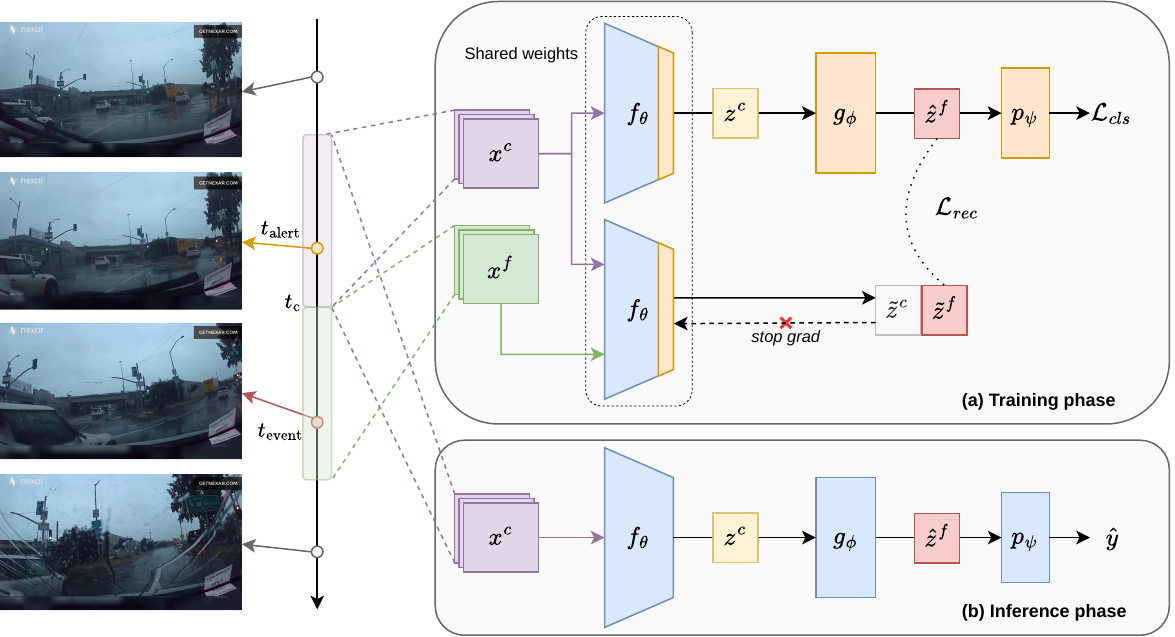}}
\caption{
Overview of FLaRA. \textbf{(a)} During training, the encoder $f_{\theta}$ processes context frames (purple segment), and the predictor forecasts future latent representations. These predictions are supervised by a reconstruction loss against target representations obtained by encoding context and future frames (green segment) jointly. Only the predicted future features are used for the classification objective. \textbf{(b)} At inference, only context frames are available, thus requiring the model to classify exclusively from predicted future representations.}
\label{fig:teaser_anticipation}
\vspace{-10pt}
\end{figure*}

Dashcam videos are becoming increasingly common, with millions of vehicles worldwide equipped with forward-facing cameras continuously recording driving scenes. This widespread availability makes them a natural and scalable data source for developing proactive safety systems applied for tasks such as viewpoint estimation~\cite{magistri2020viewestdashcam, magistri2023viewest}, anomaly detection~\cite{haresh2020anomalydashcam} and accident detection~\cite{zhou2022detectionaccident}. 
However, while modern driving safety systems excel at detecting collisions as they happen~\cite{taccari2018detection, zhou2022detectionaccident}, the ability to predict accidents before they actually occur, relying just on video data, remains an open challenge. It is crucial that the prediction be made in advance so that there is a large enough buffer to actually perform an evasive maneuver and avoid the dangerous event. Existing approaches for accident anticipation follow a common paradigm: processing observed frames and directly mapping their representations to a collision probability. Most methods operate frame-by-frame, using recurrent architectures with spatio-temporal attention to accumulate temporal evidence~\cite{chan2016anticipating,DSTA, Bao_2020}, while others apply graph neural networks alongside LSTM architectures~\cite{malawade2022gnnanticipation} or incorporate monocular depth cues for 3D scene modeling~\cite{liao2024modeling3d}.

To advance beyond this paradigm, we propose Predicting Future Latent Representations for Accident Anticipation (FLaRA), a novel predictive architecture built on top of the Video Joint-Embedding Predictive Architecture (V-JEPA~\cite{assran2025v}) that leverages video foundation models for accident anticipation.
Rather than mapping observed context directly to a classification output, as seen in a recent work by Goldshmidt et al.~\cite{goldshmidt2025badascontextawarecollision}, our approach conditions a predictor network on context frames to forecast the \textit{future representations} of the scene within a latent space. By shifting the classification task to operate on these predicted future features rather than past observations, the model effectively simulates the forthcoming progression of the scene. To ensure these forecasts accurately reflect actual future events, we introduce a joint-training paradigm. Inspired by the original self-supervised objective used in the pre-training stage of V-JEPA, our approach optimizes an auxiliary reconstruction loss between the predicted features and the ground-truth latent representations of actual \textit{future} frames, in conjunction with a standard classification loss, allowing the model to make informed inferences regarding the development of the driving scene.

Figure~\ref{fig:teaser_anticipation} shows an overview of FLaRA. During training, FLaRA observes a context window (purple) and forecasts future latent representations (green) to produce the predictions. At inference, it uses only this context window to predict the future latent representations used for classification.

We train our proposed solution on the Nexar dataset~\cite{Moura_2025_CVPR} and evaluate it under a clearly defined protocol using standard metrics (AP, AUC, mTTA). To demonstrate the robustness and generalization capabilities we further validate ours solution in out-of-domain scenarios across several established benchmarks, including DAD~\cite{chan2017dad}, DADA-2000~\cite{fang2019dada2000}, and DoTA~\cite{yao2023dota}.

Our primary contributions are summarized as follows:
\begin{itemize}
\item We introduce a novel accident anticipation framework that leverages V-JEPA2's capacity to forecast future latent representations from historical context and utilizes these predicted features to classify impending collisions.
\item We introduce a multi-task optimization strategy that combines a standard classification loss with an auxiliary latent reconstruction loss. Inspired by V-JEPA's self-supervised pre-training objective, this ensures our predicted features accurately reflect ground-truth future states.
\item We conduct extensive experiments across multiple datasets, demonstrating that our proposed method achieves highly competitive results. It matches or surpasses existing state-of-the-art baselines in predictive accuracy (AP, AUC) while maintaining an early warning capability.
\end{itemize}

\section{Related Work}
\label{sec:related_works}
Early works in accident anticipation, such as the Dynamic Spatial Attention (DSA) model \cite{chan2016anticipating}, utilized LSTM networks to evaluate candidate objects and generate frame-by-frame future accident probabilities. Subsequent developments expanded upon these foundations by incorporating novel modules to model object interactions. For instance, UString \cite{Bao_2020} leverages Recurrent Neural Network and Graph Convolutional Network architectures to capture complex spatio-temporal relationships. A key contribution of UString is its adaptive loss function, designed to facilitate early hazard prediction by dynamically adjusting loss weights based on temporal proximity to the collision. Following this, DSTA \cite{DSTA} employed Transformer-based architectures equipped with dynamic spatial-temporal attention mechanisms. This design enables the model to effectively isolate highly relevant spatial regions while simultaneously tracking their temporal evolution across video sequences. More recently, Pjetri et al.~\cite{pjetri2025selfsupervisedanticipation} proposed a self-supervised approach that exploits a non-decreasing danger assumption to design a loss function that does not require hand-crafted temporal annotations.

Shifting toward video foundation models, recent work has explored V-JEPA2~\cite{assran2025v}, a self-supervised architecture that learns by predicting latent representations of masked spatio-temporal regions. Adapting this powerful encoder for accident anticipation, Goldshmidt et al.~\cite{goldshmidt2025badascontextawarecollision} proposed BADAS. This method fine-tunes V-JEPA2 alongside a dedicated prediction head, yet it continues to base its collision predictions strictly on the representations of the observed context window.

An alternative line of work in action anticipation instead proposes to first forecast future visual representations and classify on those. Vondrick et al.~\cite{Vondrick_2016_CVPR} is one of the first methods to do this. They trained a network to predict future CNN representations from the current frame, then training a classifier on the predicted output. Gammulle et al.~\cite{Gammulle_2019_ICCV} further extended this approach through a recurrent GAN framework that jointly optimizes future visual and temporal feature synthesis alongside a classification objective. Assran et al.~\cite{assran2025v} also build on the pretrained V-JEPA2, applying it to action anticipation. Given currently observed frames, the frozen V-JEPA2 predictor produces future latent representations. These predicted features are concatenated with the encoded present features and passed to an attentive probe, which is the only component trained for the downstream task, using only classification loss.
Although this predict-then-classify paradigm has been explored in action anticipation, to our knowledge, it has not been applied to the accident anticipation domain.

\section{Methodology}
\label{sec:methodology}

In this section we formalize our approach. We first define the problem setup (Section~\ref{subsec:problem_setup}), then describe the architecture and training procedure, including the joint reconstruction-classification objective (Section~\ref{subsec:training_architecture}), and finally detail the inference scheme based on predicted future representations (Section~\ref{subsec:inference}).

\subsection{Problem Setup}
\label{subsec:problem_setup}

Given a video sequence, the goal of accident anticipation is to predict whether a safety-critical event will occur using only the visual information that precedes it. 

Let $V=(v_1, v_2, \dots, v_T)$ denote a video sequence of $T$ frames. Let $t_{\text{event}}$ represent the timestamp of a safety-critical event, and let $k_{\text{event}}$ be the corresponding discrete frame index. The objective is to learn a model $\mathcal{M}$ that predicts the label $y \in \{0, 1\}$ using an observed context window up to frame $k_c$, where $k_c < k_{\text{event}}$. The prediction is given by:
\begin{equation}
    \hat{y} = \mathcal{M}(v_1, v_2, \dots, v_{k_c}).
\end{equation}
This anticipates the event with a lead time of $k_{\text{event}} - k_c$ frames (or $t_{\text{event}} - t_c$ seconds, where $t_c$ is the timestamp of frame $k_c$).

\subsection{Architecture and Training}
\label{subsec:training_architecture}
Our approach is based on the idea of leveraging predicted future representations for classification~\cite{Vondrick_2016_CVPR}. Instead of using the features of the observed frames directly, a predictor network conditioned only on the context window produces feature-level forecasts. 
These predicted features are then used as inputs for the classification task. To ensure the predictor learns meaningful future representations and adapts to the new distribution, we supervise it during training with an auxiliary reconstruction loss where the targets are the encoder's representations of the ground-truth future frames.

In particular, our model $\mathcal{M}$ builds on V-JEPA2 and comprises three main components: a video encoder $f_\theta$, a future predictor $g_\phi$ and an attentive probe $p_\psi$~\cite{lee2019settransformer,assran2025v}. The probe is composed of four transformer blocks. The first three are self-attention layers that refine the input token representations. The final block replaces self-attention with a cross-attention layer where a single learnable query token attends to the refined features, producing a pooled representation. This is then passed to a linear layer for classification.

During training (see Figure~\ref{fig:teaser_anticipation}(a)) the model receives both context frames and future frames. Let $x^c = (v_1, \dots, v_{k_c})$ denote the sequence of context frames, and let $x^f = (v_{k_c+1}, \dots, v_{k_c+L})$ denote the sequence of future frames, which spans a fixed length of $L$ frames immediately following the context window. The context frames are sampled such that the window ends at $t_{\text{event}} - t_c > 0$ seconds prior to the potential event, matching our anticipation time window.

The encoder $f_\theta$ first processes both context and future frames jointly to establish targets, producing $\tilde{z} = f_{\theta}([x^c, x^f]) = [\tilde{z}^c, \tilde{z}^f] \in \mathbb{R}^{(N_c+N_f) \times D}$, where $N_c$ and $N_f$ are the number of context and future tokens respectively, $D$ is the encoder embedding dimension, $\tilde{z}^c$ is the feature representation of the context, and $\tilde{z}^f$ is the representation of the future. The encoder also processes the context frames $x^c$ without access to the future frames, producing $z^c = f_{\theta}(x^c) \in \mathbb{R}^{N_c \times D}$. Then, the predictor $g_{\phi}$ maps the context features $z^c$ to the predicted future representations $\hat{z}^f = g_{\phi}(z^c) \in \mathbb{R}^{N_f \times D}$.

To guide the predictor to produce meaningful future representation, we compute an auxiliary reconstruction loss:
\begin{equation}
\label{eq:rec_loss}
    \mathcal{L}_{rec} = \frac{1}{N_f} \sum_{i=1}^{N_f} \ell(\hat{z}_i^f, \mathrm{sg}(\tilde{z}_i^f)),
\end{equation}
where $\ell$ denotes the Smooth-L1 loss, $\mathrm{sg}(\cdot)$ is the stop-gradient operator, and $\hat{z}_i^f$ and $\tilde{z}_i^f$ denote the $i$-th token of the predicted and target future representations, respectively.

Simultaneously, we compute the classification loss using the cross-entropy objective:
\begin{equation}
    \mathcal{L}_{cls} = \ell_{CE}(\hat{y}, y),
\end{equation}
where $\hat{y} = p_{\psi}(\hat{z}^f)$ is the prediction derived from passing the forecasted future representations through the attentive probe. 

The overall training objective is formulated as:
\begin{equation}
    \mathcal{L} = \mathcal{L}_{cls} + \lambda \mathcal{L}_{rec},
\end{equation}
where $\lambda \in \mathbb{R}$ is a hyperparameter modulating the influence of the reconstruction loss.

The reconstruction loss in Eq.~\ref{eq:rec_loss} adopts the V-JEPA2 self-supervised pretraining objective but differs in application.
While V-JEPA2 reconstructs randomly masked spatio-temporal patches within a single clip, FLaRA uses as targets the latent representations of subsequent temporal frames.
Furthermore, $\mathcal{L_{\text{rec}}}$ serves as an auxiliary signal alongside the classification loss, while the original V-JEPA2 pretraining has only the reconstruction objective.
Mapping targets to future frame features aligns the model with the requirements of accident anticipation by anchoring outputs to future scene dynamics.

FLaRA represents the first attempt to adapt the V-JEPA2 predictor for accident anticipation by utilizing a reconstruction signal and performing classification exclusively on predicted future latent representations.

\subsection{Inference}
\label{subsec:inference}
At inference time (see Figure~\ref{fig:teaser_anticipation}(b)), only the context frames $x^c$ are available. The encoder extracts the context features $z^c = f_{\theta}(x^c)$, which the predictor then uses to forecast the future representations $\hat{z}^f = g_{\phi}(z^c)$. The final classification is performed by computing the prediction $\hat{y} = p_{\psi}(\hat{z}^f)$. In this way the real future frames $x^f$ are \textit{never} accessed.

\section{Experimental Results}
\label{sec:experimental_results}

We begin by describing the datasets and implementation details, followed by the evaluation protocol and metrics. We then compare our method against prior work and conclude with ablation studies and analysis.

\subsection{Datasets}
\label{subsec:datasets}

We train our model on the training split of the Nexar dataset~\cite{Moura_2025_CVPR}, which consists of 1,500 samples evenly balanced between positive and negative clips. 

Each positive clip is annotated with two timestamps: the earliest moment of recognizable danger $t_{\text{alert}}$ and the precise time of the safety critical event $t_{\text{event}}$, which can be either a collision or a near-collision. We use $t_{\text{event}}$ to ensure that the sampled context window never includes frames of the actual event.
We evaluate our approach on the Nexar test split, comprising 1,344 clips (672 positive and 672 negative). Each positive clip is temporally cropped prior to the safety-critical event using a variable offset $\Delta t \in \{0.5, 1.0, 1.5\}$ seconds.

To evaluate cross-dataset generalization, we further evaluate our model on three additional benchmarks: DAD~\cite{chan2017dad}, DADA-2000~\cite{fang2019dada2000} and DoTA~\cite{yao2023dota} datasets, using the subsets and re-annotations defined by~\cite{goldshmidt2025badascontextawarecollision}.
These subsets retain only ego-vehicle collisions and include manually annotated $t_{\text{alert}}$ timestamps, which are absent or inconsistent from the original datasets.
As summarized in Table~\ref{tab:eval_datasets}, these subsets are heavily imbalanced. We also report mean $t_{\text{event}}$ and mean $t_{\text{event}} - t_{\text{alert}}$ computed for each dataset by averaging across positive samples.
Furthermore, because the original DADA-2000 and DoTA datasets lack negative samples, following~\cite{goldshmidt2025badascontextawarecollision} we construct negatives by extracting the first 4.0 seconds from positive clips in which the alert timestamp ($t_{\text{alert}}$) occurs at least 4.5 seconds after the start of the video, guaranteeing that the sampled segment captures only normal driving dynamics and no temporal overlap with the safety-critical event.

\begin{table}
    \centering
    \caption{Benchmark statistics. For DAD, DADA-2000, and DoTA, we use the specific subsets defined by~\cite{goldshmidt2025badascontextawarecollision}.}
    \resizebox{\columnwidth}{!}{%
    \begin{tabular}{lcc ccc}
        \toprule
        \textbf{Dataset} & \textbf{Positives} & \textbf{Negatives} & \textbf{Mean $t_{\text{event}}$} & \textbf{Mean $t_{\text{event}} - t_{\text{alert}}$} \\
        \midrule
        Nexar (test)  & 672 & 672 & 10.9 & 1.9 \\
        DAD           & 13  & 301 & \phantom{0}3.0 & 1.2 \\
        DADA-2000     & 75  & 38 & \phantom{0}6.8 & 1.8 \\
        DoTA          & 327 & 40 & \phantom{0}5.1 & 2.0 \\
        \bottomrule
    \end{tabular}
    \label{tab:eval_datasets}
}
\vspace{-10pt}
\end{table}

\subsection{Implementation Details}
\label{subsec:implementation_details}

We initialize our model using a V-JEPA2 ViT-L backbone pretrained on the Something-Something v2 dataset~\cite{goyal2017something}, which provides the initial weights for the encoder, predictor, and attentive probe. During fine-tuning, we freeze the encoder except for its final block. Only this final encoder block, the predictor, the attentive probe and the newly instantiated classifier are updated, resulting in 84M trainable parameters (see Table~\ref{tab:nexar_results}).

Both the context clip $x^c$ and future clip $x^f$ consist of 16 frames sampled at 8 fps with a spatial resolution of $256 \times 256$. The future frames are sampled contiguously, starting immediately after the end of the context window (see Figure~\ref{fig:teaser_anticipation}). For positive samples, the context clip is extracted such that its last frame precedes the collision by a random offset sampled uniformly from $[0,\gamma]$,
where $\gamma = t_{\text{event}} - t_{\text{alert}}$. This ensures the model is exposed to clips at varying temporal distances from the event, spanning from the moment the alert was triggered up to the collision boundary.

We train the network for 30 epochs using a batch size of 4, using AdamW with a weight decay of $1 \times 10^{-2}$ and using a peak learning rate of $2.5 \times 10^{-5}$. We also adopt a cosine annealing schedule warmup. We sweep the reconstruction loss weight over $\lambda \in \{1, 2, 10, 100\}$ and use $\lambda = 10$ in all experiments. Reported results are averaged over three runs and all experiments are run on a single RTX 4090 GPU.

\subsection{Evaluation Protocol}
\label{subsec:evaluation_protocol}
At inference, we use our model in a sliding window fashion over the input video. Each window consists of 16 frames sampled at 8 fps, mirroring the training configuration. The window has a stride of 8 video frames, resulting in overlapping windows. For positive videos, only windows whose end time falls before the annotated time of event $t_{\text{event}}$ are considered valid, ensuring the model cannot observe the safety-critical event itself.
For negative videos, the full video is evaluated.

Following standard evaluation protocol metrics~\cite{Bao_2020,DSTA}, we report Area Under the ROC Curve (AUC), Average Precision (AP) and mean Time-to-Accident (mTTA).
For AUC, the score is aggregated per video by selecting the maximum score among all the sliding windows.
For AP and mTTA, we sweep detection thresholds $\sigma$ by considering a positive video detected if any valid window exceeds the threshold.
For positive samples, the Time-to-Accident (TTA) at a given threshold $\sigma$ is defined as:
\begin{equation}
    \text{TTA}(\sigma) = t_{\text{event}} - t_{\text{det}}(\sigma),
\end{equation}
where $t_{\text{det}}(\sigma)$ denotes the \textit{end} timestamp of the earliest window exceeding the threshold.
We use the final frame of the temporal window, as the model requires the complete 16-frame sequence to generate an inference.
We then compute the mean Time-to-Accident (mTTA) by averaging TTA values over true positive videos detected at each threshold.
The final metric is thus obtained by averaging these results across all recall operating points.

We evaluate our method against UString~\cite{Bao_2020}, DSTA~\cite{DSTA}, both trained on the CCD dataset~\cite{Bao_2020} and BADAS~\cite{goldshmidt2025badascontextawarecollision}, trained on Nexar, using the publicly available checkpoints.
All clip-based methods (Ours, BADAS) are evaluated within this same sliding window framework. UString and DSTA are per-frame recurrent models that produce a prediction at each timestep using all previously observed frames. For these models, we extract features at native video FPS and run the temporal model over the full sequence, yielding one prediction per frame. These per-frame scores are then fed into the same metric computation pipeline without any windowing or temporal resampling.

\subsection{Comparative Performance Evaluation}
\label{subsec:comparative_performance_evaluation}

Table~\ref{tab:nexar_results} presents the performance of our method on the Nexar test set compared to prior approaches. Our model achieves the highest AP ($86.6$) and AUC ($87.3$), outperforming the previous best performing model BADAS while using roughly $4\times$ fewer trainable parameters (84M vs 332M). In terms of mTTA values, we are comparable to BADAS, indicating that the improvement on AP and AUC did not come at the cost of later warnings.

Table~\ref{tab:cross_domain_results} reports cross-dataset generalization, where all methods are evaluated on DAD, DADA-2000 and DoTA.
On DAD, FLaRA achieves a substantial improvement over BADAS ($+18.4$ AP), indicating a stronger generalization to unseen driving scenarios, while on DADA-2000 and DoTA our model still performs competitively. These results are achieved without any domain-specific fine-tuning, demonstrating that the predicted future representations learned on Nexar transfer to other accident anticipation benchmarks.

Across datasets, mTTA values should be interpreted in light of the video characteristics we reported in Table~\ref{tab:eval_datasets}. Both FLaRA and BADAS require a minimum window of 2 seconds (16 frames at 8 fps), which constrains how early detection can occur on short clips. On datasets like DAD, where mean $t_{\text{event}}$ is just $3.0$ seconds, there is very little margin, explaining the low mTTA values across all clip-based methods. Conversely, on the Nexar dataset, $t_{\text{event}}$ is 10.9 seconds, allowing the model to start observing context windows further away from the actual event and potentially achieve earlier detections. Moreover, the difference $t_{\text{event}} - t_{\text{alert}}$ averages 1.9 seconds, meaning that our mTTA of 3.744 seconds indicates detection well before visual cues of safety-critical events are recognizable by a human observer.

\begin{table}
\caption{Accident anticipation performance on Nexar.
}
\label{tab:nexar_results}
    \centering
    \begin{tabular}{l r  ccc}
        \toprule
        \textbf{Method} & \textbf{N. Params} & \textbf{AP} $\uparrow$ & \textbf{AUC} $\uparrow$ & \textbf{mTTA} \\
          \cmidrule(lr){1-1}
          \cmidrule(lr){2-2}
          \cmidrule(lr){3-5}
         DSTA & 4.8M & 50.0 & 45.5 & 10.216 \\
         UString & 1.9M & 49.6 & 41.0 & 10.263 \\
         BADAS & 332M & 85.2 & 85.1 & \phantom{0}3.712 \\
          \cmidrule(lr){1-1}
          \cmidrule(lr){2-2}
          \cmidrule(lr){3-5}
         \textbf{FLaRA} & 84M & \textbf{86.6} & \textbf{87.3}& \phantom{0}3.744 \\
        \bottomrule
    \end{tabular}
\end{table}

\begin{table}
\caption{Generalization performance on unseen accident anticipation benchmarks.}
\label{tab:cross_domain_results}
    \centering
    \resizebox{0.65\columnwidth}{!}{%
    \begin{tabular}{l  l  ccc}
        \toprule
        \textbf{Dataset} & \textbf{Method} & \textbf{AP} $\uparrow$ & \textbf{AUC} $\uparrow$ & \textbf{mTTA} \\
          \cmidrule(lr){1-1}
          \cmidrule(lr){2-2}
          \cmidrule(lr){3-5}
        \multirow{5}{*}{\textbf{DAD}} 
        & DSTA & \phantom{0}4.7 & 61.2 & 2.832 \\
        & UString & \phantom{0}5.5 & 65.6 & 2.694 \\
        & BADAS & 48.9 & 92.5 & 0.782 \\
          \cmidrule(lr){2-2}
          \cmidrule(lr){3-5}
        & \textbf{FLaRA} & \textbf{67.3} & \textbf{93.6} & 0.663 \\
        \midrule
        \multirow{5}{*}{\textbf{DADA}} 
        & DSTA & 66.4 & 52.5 & 6.552 \\
        & UString & 66.4 & 58.1 & 6.598 \\
        & BADAS & \textbf{98.0} & \textbf{95.0} & 1.601 \\
          \cmidrule(lr){2-2}
          \cmidrule(lr){3-5}
        & \textbf{FLaRA} & 97.3 & 93.2 & 1.595 \\
        \midrule
        \multirow{5}{*}{\textbf{DoTA}} 
        & DSTA & 89.1 & 48.1 & 4.759 \\
        & UString & 88.5 & 50.8 & 4.732 \\
        & BADAS & 98.7 & \textbf{90.6} & 1.562 \\
        \cmidrule(lr){2-2}
        \cmidrule(lr){3-5}
        & \textbf{FLaRA} & \textbf{98.8} & 90.0 & 1.649 \\
        \bottomrule
    \end{tabular}
    }%
    \vspace{-10pt}
\end{table}

\subsection{Ablation Study}
\label{subsec:ablation_and_analysis}

Table~\ref{tab:ablations} presents an ablation study isolating the contribution of each component. \textit{Frozen Backbone} keeps the encoder $f_{\theta}$ and predictor $g_{\phi}$ frozen, training only the attentive probe $p_{\psi}$ on the classification objective. Since $g_{\phi}$ receives no training signal, it relies entirely on V-JEPA2's pretrained representations. The \textit{Fine-tuned w/o $\mathcal{L}_{\text{rec}}$} matches our full training setup, but removes $\mathcal{L}_{\text{rec}}$, meaning $g_{\phi}$ is guided only by the classification objective.
Comparing these variants demonstrates that fine-tuning $g_{\phi}$ and last layer of $f_{\theta}$ results in a substantial improvement, especially on Nexar ($+5.1$ AP) and DADA-2000 ($+5.4$ AP). 
However, without the reconstruction objective, $g_{\phi}$ is guided solely by the classification gradient, without any guarantee that its outputs correspond to meaningful future representations.

Introducing $\mathcal{L_{\text{rec}}}$ addresses this by guiding the predicted features closer to the latent dynamics of the actual future scene. This drives consistent gains across all benchmarks, most notably on DAD, where AP grows from $25.6$ to $67.3$. Because the DAD subset exhibits a severe class imbalance (only 13 accidents in 314 clips) that closely mirrors the real-world rarity of collisions, this confirms that training on semantically coherent future predictions is crucial for generalizing to \textit{realistic}, rare-event domains.

In the \textit{Encoder Only} ablation we remove $g_{\phi}$ entirely and classify directly from the fine-tuned encoder features, effectively using only current frames information. 
This ablation mirrors BADAS's training regime, fine-tuning the encoder directly, while retaining our reduced parameter count. While on Nexar this variant has closer performance to our full model, cross-dataset performance drops considerably, confirming that the predictor, guided by $\mathcal{L_{\text{rec}}}$, learns future representations that generalize well to other unseen distributions, improving with respect to the \textit{Encoder Only} training regime.

Finally, the \textit{FLaRA + $z^c$} variant retains our full training objective but classifies using the concatenation of context ($z^c$) and predicted future ($\hat{z}^f$) features. This degrades performance on Nexar (-1.5 AP) and DAD, suggesting that present features hinder classification by introducing redundant information that interferes with the model's discriminative capabilities. While performance on DADA-2000 and DoTA remains comparable, classifying exclusively from predicted future features proves to be the most effective design for accident anticipation.

\begin{table}
\caption{Ablation study on the effect of fine-tuning, reconstruction loss and features used for classification.}
\label{tab:ablations}
    \centering
    \resizebox{0.90\columnwidth}{!}{%
    \begin{tabular}{l ccc ccc}
        \toprule
        & \multicolumn{3}{c}{\textbf{Nexar}} & \multicolumn{3}{c}{\textbf{Dad}} \\
        \cmidrule(lr){2-4} \cmidrule(lr){5-7}
        \textbf{Method} & \textbf{AP} $\uparrow$ & \textbf{AUC} $\uparrow$ & \textbf{mTTA} & \textbf{AP} $\uparrow$ & \textbf{AUC} $\uparrow$ & \textbf{mTTA} \\
        \midrule
        Frozen Backbone & 80.4 & 81.8 & 4.688 & 22.5 & 77.6 & 0.651 \\
        Fine-tuned w/o $\mathcal{L}_{\text{rec}}$ & 85.5 & 87.0 & 3.882 & 25.6 & 88.0 & 0.788 \\
        Encoder Only & 86.2 & 87.3 & 3.867 & 35.8 & 80.7 & 0.713 \\
        FLaRA + $z^c$ & 85.1 & 86.0 & 3.570 & 58.4 & \textbf{93.7} & 0.654 \\
        \cmidrule(lr){1-1} \cmidrule(lr){2-4} \cmidrule(lr){5-7}
        \textbf{FLaRA} & \textbf{86.6} & \textbf{87.3} & 3.744 & \textbf{67.3} & 93.6 & 0.663 \\
        \midrule
        & \multicolumn{3}{c}{\textbf{DADA-2000}} & \multicolumn{3}{c}{\textbf{Dota}} \\
        \cmidrule(lr){2-4} \cmidrule(lr){5-7}
        \textbf{Method} & \textbf{AP} $\uparrow$ & \textbf{AUC} $\uparrow$ & \textbf{mTTA} & \textbf{AP} $\uparrow$ & \textbf{AUC} $\uparrow$ & \textbf{mTTA} \\
        \midrule
        Frozen Backbone & 90.2 & 81.9 & 1.774 & 97.9 & 85.0 & 1.568 \\
        Fine-tuned w/o $\mathcal{L}_{\text{rec}}$ & 95.6 & 90.1 & 2.056 & 98.7 & 89.4 & 1.783 \\
        Encoder Only & 95.3 & 89.7 & 1.748 & 98.4 & 87.4 & 1.619 \\
        FLaRA + $z^c$ & 97.0 & 92.8 & 1.551 & \textbf{98.8} & \textbf{90.0} & 1.511 \\
        \cmidrule(lr){1-1} \cmidrule(lr){2-4} \cmidrule(lr){5-7}
        \textbf{FLaRA} & \textbf{97.3} & \textbf{93.2} & 1.595 & \textbf{98.8} & \textbf{90.0} & 1.649 \\
        \bottomrule
    \end{tabular}%
    }
\vspace{-10pt}
\end{table}

\subsection{Anticipation Horizon Analysis}
\begin{figure}
\centerline{\includegraphics[width=0.3\textwidth]{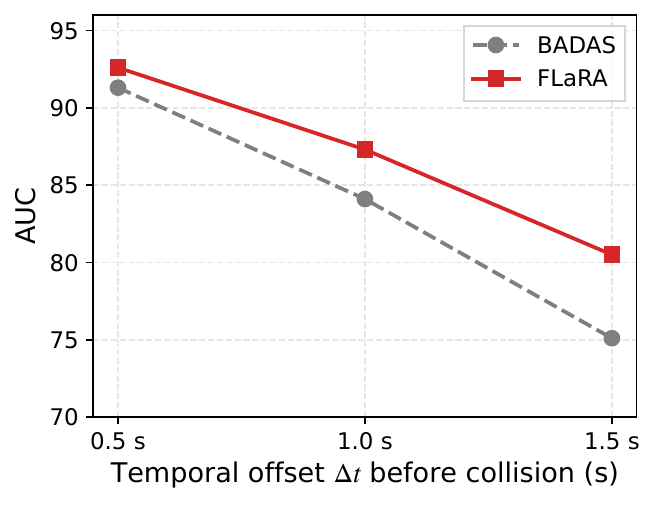}}
\caption{AUC on the Nexar test set as a function of the temporal offset before the collision event. Videos are grouped by their annotated prediction horizons of $\Delta t \in \{0.5, 1.0, 1.5\}.$
}
\label{fig:temporal_offset}
\vspace{-10pt}
\end{figure}

The Nexar test set contains positive videos cropped at different temporal distances from the safety-critical event ($\Delta t \in \{0.5, 1.0, 1.5\}$ seconds). Figure~\ref{fig:temporal_offset} reports the AUC on positive videos filtered by offset, comparing our method to BADAS. As expected, performance degrades for both methods as temporal distance increases, since earlier crops contain fewer visual cues of the impending collision. However, FLaRA consistently outperforms BADAS across all offsets, with the performance gap widening at larger distances (+1.3 AUC at $\Delta t = 0.5$ compared to +5.4 AUC at $\Delta t = 1.5$). This demonstrates that predicting future representations is particularly \textit{beneficial} with larger horizons, where the observed context alone provides limited discriminative information and the model must rely on forecasting the scene's evolution.

\section{Conclusions and Limitations}
\label{sec:conclusions}

In this work we showed that forecasting future latent representations and using them for classification, rather than relying on observed features, consistently improves accident anticipation performance. The joint reconstruction-classification objective we introduced anchors the role of the V-JEPA2 predictor to output actual future scene latent representations and allows effective adaptation of the backbone and predictor to the target domain. Our framework achieves competitive or superior results across four benchmarks with four times fewer trainable parameters than the previous state-of-the-art: the largest gains are achieved on DAD, whose class distribution most closely reflects real-world distributions, and on the improvement of detection capabilities at higher temporal offsets, as shown in Figure~\ref{fig:temporal_offset}.
However, our approach comes with some limitations. While the detection quality of the model improves, the anticipation time remains comparable to BADAS, our strongest competitor, and improving mTTA without sacrificing precision remains an open challenge. Also, the predictor operates with a fixed temporal horizon. As future work, introducing an adaptive prediction range or multiple predictions could better reflect and improve performance in real-world applications, by allowing the model to predict further in the future when far from the event. In addition, architectural scaling and hardware-specific optimizations to reduce the inference cost represents an interesting direction for real-time edge deployment.

\bibliographystyle{IEEEtran}

\bibliography{main_bib}

\end{document}